# Initial Comparison of Linguistic Networks Measures for Parallel Texts

Kristina Ban, Ana Meštrović , Sanda Martinčić-Ipšić
Department of Informatics
University of Rijeka
Radmile Matejčić 2, 51000 Rijeka, Croatia
kristina.ban89@gmail.com, amestrovic@uniri.hr, smarti@uniri.hr

**Abstract:** *In this paper we compared the properties of linguistic networks for Croatian, English and Italian languages. We constructed co-occurrence networks from parallel text corpora, consisting of the translations of five books in the three languages. We generated an Erdös-Rényi random graph with the same number of nodes and links, which enabled the comparison with linguistic co-occurrence networks, showing small-world properties. Furthermore, the comparison of Croatian, English and Italian linguistic networks showed that, besides expected commonalities of networks, there are also certain differences. The networks' measures across the three studied languages differ particularly in the average path length. The results indicate that size of the corpus and anomalies in text affect the network structure.*

## 1. Introduction

Network analysis nowadays exhibits a growing popularity because it provides a way to analyse real complex systems. Language is an example of a complex system and in the last decade it has been the subject of many network based studies, highlighting the field of linguistic networks. Various linguistic networks can be analysed such as syntax networks [2-4], semantics' networks [5], phonological networks [6-8], syllable networks [14-15], word co-occurrence networks [1,8].

The focus of the research in linguistic networks has shifted from one language to multiple languages. The work in [10] examines structural differences in Chinese and English by comparing the intensity and density of the connection in networks. In [9] the network properties of the English and German Wikipedia are compared. The paper by Liu and Jin [11] studied language networks on multilingual parallel texts of 15 languages. One of the 12 Slavic languages was Croatian. Network parameters were used for the hierarchical classification of the languages.

Besides multiple language studies (language differentiation and classification) the research community's attention is also focused on the genre of literature or author detection, based on the analysis of complex networks. The authors in [21] examine the correlation between the network properties and author characteristics in terms of clustering the coefficient, in and out degree, degree distribution and component dynamics. The corpus used included over 40 books by eight authors in English. The work [16] investigates the properties of the writing style of five Persian authors in 36 books. The network derived measures: degree distribution and power law α-exponent were used for authorship identification.

Our research is an initial attempt at the analysis of parallel corpora of Croatian, Italian and English literature. We examined the comparative network properties of three languages in terms of language and book differentiation. The parallel nature of the corpus, consisting of the translations of five books in three languages, enables the opportunity to compare network properties across languages and to check the translation consistency on the book level.

Section 2 of the paper presents key measures of complex networks. Section 3 discusses the experiments set up and in Section 4 the results are shown. The paper concludes with the discussion and further research plans.

## 2. Methodology

Every network is constructed of nodes $N$ and links $K$. The degree $k_i$ of a node $i$ is the number of connections that the node has. The average degree of the network is:
$$<k> = \frac{2K}{N}.$$
In the case of directed networks, the in-degree and out-degree should be considered, and in this case the factor two is absent.

For every two connected nodes $i$ and $j$ the number of links lying on the path between them is denoted as $d_{ij}$, therefore the average distance of a node $i$ from all other nodes is:
$$d_i = \frac{\sum_j d_{ij}}{N}.$$
The average path distance $L$ is an average value of $d_i$ of all nodes:
$$L = \frac{\sum_i d_i}{N},$$
and the maximum distance between two nodes in the network is diameter, $D$:
$$D = \max_i d_i.$$
The clustering coefficient $c_i$ of a node $i$ is described as a probability of the presence of a link between any two neighbours of a node. It is calculated as a ratio between the number of links $E_i$ that actually exist amongst these and the total possible number:
$$c_i = \frac{2E_i}{k(k-1)}.$$
The average clustering of a network $C$ is the average value of the clustering coefficient of all the nodes:
$$C = \frac{\sum_i c_i}{N}.$$

One of the commonly examined properties of real world networks are small-world properties [1]. The network is a small-world if its average shortest path length $L \approx L_{ER}$ and its clustering coefficient $C \ll C_{ER}$ where $L_{ER}$ is the average shortest path length and $C_{ER}$ is the clustering coefficient of an Erdös-Rényi (ER) random graph with the same number of nodes and links [22].

## 3. Experiments

### 3.1. Data

We prepared a twofold balanced corpus: parallel translations of five books in Croatian, Italian and English. Each book was originally written in one and translated to the other two languages. We took care that for each language at least one native book is present and the length of the books varies from short to long (Table 1).

The text was cleared of the index of contents, the author's biography and page numbers. Afterwards the corpus was tokenized, the punctuation marks, special characters, and stopwords were removed and inflected word forms were lemmatized. For Croatian we used the stopwords list of 2922 words, for English 341 words and Italian 371 words. Table 1 shows the number of words with and without stopwords per book depending on the language. For Croatian we used the Croatian Lemmatization Server [20] for Italian and English TreeTagger [19].

Table 1. The total number of words in the books with and without stopwords by book and by language. The Croatian books show a smaller number of words but after the removal of the stopwords the total number of words is higher than the Italian and English

|  | Book | With stopwords | Without stopwords |
|---|---|---|---|
| English | B1-EN | 47684 | 16372 |
| English | B2-EN | 147537 | 56525 |
| English | B3-EN | 27299 | 10120 |
| English | B4-EN | 235245 | 89245 |
| English | B5-EN | 204517 | 76476 |
| Italian | B1-IT | 48487 | 33657 |
| Italian | B2-IT | 156325 | 115855 |
| Italian | B3-IT | 25523 | 20136 |
| Italian | B4-IT | 235207 | 183435 |
| Italian | B5-IT | 213147 | 157878 |
| Croatian | B1-HR | 44433 | 18627 |
| Croatian | B2-HR | 125997 | 59293 |
| Croatian | B3-HR | 24507 | 10973 |
| Croatian | B4-HR | 217987 | 100308 |
| Croatian | B5-HR | 198188 | 90299 |

### 3.2. Networks construction from books

We constructed a co-occurrence network for each book: 15 directed and 15 undirected. Words are represented as nodes and linked if they appear as adjacent words in the text. Additionally, we also generated an ER random graph with the same number of nodes and links for each network.

We used the Python programming language with its module NLTK [20] for text processing, the NetworkX module [12] for the construction of the networks and analysis, and Gephi software [13] for the manipulation of the networks and visualization.

## 4. Results

As shown in Tables 2 and 3, co-occurrence networks based on parallel texts share common properties: a small average path length $L$ and diameter $D$ and a high clustering coefficient $C$ in comparison with its associated ER graph. The difference between the clustering coefficient of the linguistic and the random networks varies from the minimum $C_{DIR} \approx 29 C_{ER}$ to the maximum $C_{DIR} \approx 148 C_{ER}$. The linguistic networks for all three languages have small-world properties. Another shared property of the undirected networks is a higher $C$ and smaller $L$ and $D$, contrary to the same measures of directed network of the same book. This means that undirected networks are closer to the small-world networks, which is an expected result. However, there is one exception with the results for book B5 the diameter of which increased in the undirected network.

**Table 2. The result for the directed networks of five books in three languages: N - number of nodes, <k> average node degree, $C_{DIR}$ – clustering coefficient, $L_{DIR}$ – average path length, $D_{DIR}$ – diameter. $C_{ER}$ - clustering coefficient, $L_{ER}$ average path length and $D_{ER}$ diameter of ER random graph.**

|  |  | $N$ | $<k>$ | Directed | | | Erdős–Rényi | | |
|---|---|---|---|---|---|---|---|---|---|
|  |  |  |  | $C_{DIR}$ | $L_{DIR}$ | $D_{DIR}$ | $C_{ER}$ | $L_{ER}$ | $D_{ER}$ |
| English | B1-EN | 2389 | 5.40 | 0.070 | 3.33 | 10 | 0.00228 | 4.60 | 15 |
|  | B2-EN | 7322 | 6.50 | 0.054 | 3.56 | 13 | 0.00089 | 5.34 | 19 |
|  | B3-EN | 1798 | 4.38 | 0.076 | 3.23 | 10 | 0.00247 | 4.53 | 14 |
|  | B4-EN | 12126 | 5.87 | 0.072 | 3.52 | 12 | 0.00049 | 5.93 | 20 |
|  | B5-EN | 10027 | 6.38 | 0.051 | 3.64 | 14 | 0.00064 | 5.59 | 20 |
| Italian | B1-IT | 3858 | 4.28 | 0.052 | 3.51 | 13 | 0.00111 | 5.06 | 17 |
|  | B2-IT | 9120 | 6.45 | 0.044 | 3.64 | 13 | 0.00071 | 5.51 | 21 |
|  | B3-IT | 2269 | 4.30 | 0.068 | 3.32 | 10 | 0.00191 | 4.65 | 14 |
|  | B4-IT | 14009 | 6.34 | 0.047 | 3.62 | 14 | 0.00045 | 5.95 | 22 |
|  | B5-IT | 13403 | 5.86 | 0.044 | 3.65 | 14 | 0.00044 | 6.01 | 20 |
| Croatian | B1-HR | 4155 | 3.74 | 0.047 | 3.65 | 12 | 0.00090 | 5.09 | 16 |
|  | B2-HR | 12610 | 4.23 | 0.034 | 3.92 | 13 | 0.00033 | 5.93 | 21 |
|  | B3-HR | 2970 | 3.23 | 0.049 | 3.51 | 11 | 0.00110 | 4.69 | 15 |
|  | B4-HR | 15256 | 5.40 | 0.051 | 3.74 | 13 | 0.00036 | 6.20 | 20 |
|  | B5-HR | 15985 | 4.91 | 0.038 | 3.87 | 14 | 0.00031 | 6.25 | 21 |

**Table 3. The result for the undirected networks of five books in three languages: N - number of nodes, <k> average node degree, $C_{DIR}$ – clustering coefficient, $L_{DIR}$ – average path length, $D_{DIR}$ – diameter. $C_{ER}$ - clustering coefficient, $L_{ER}$ average path length and $D_{ER}$ diameter of ER random graph.**

|  |  | $N$ | $<k>$ | Undirected | | | Erdős–Rényi | | |
|---|---|---|---|---|---|---|---|---|---|
|  |  |  |  | $C_{UNDIR}$ | $L_{UNDIR}$ | $D_{UNDIR}$ | $C_{ER}$ | $L_{ER}$ | $D_{ER}$ |
| English | B1-EN | 2389 | 10.8 | 0.145 | 3.32 | 8 | 0.005 | 3.52 | 6 |
|  | B2-EN | 7322 | 13 | 0.109 | 3.36 | 8 | 0.002 | 3.74 | 6 |
|  | B3-EN | 1798 | 8.76 | 0.155 | 3.30 | 8 | 0.004 | 3.67 | 6 |
|  | B4-EN | 12126 | 11.74 | 0.144 | 3.52 | 8 | 0.001 | 4.07 | 7 |
|  | B5-EN | 10027 | 12.76 | 0.103 | 3.60 | 23 | 0.001 | 4.00 | 7 |
| Italian | B1-IT | 3858 | 8.56 | 0.108 | 3.45 | 9 | 0.003 | 4.08 | 7 |
|  | B2-IT | 9120 | 12.9 | 0.088 | 3.35 | 11 | 0.001 | 3.83 | 6 |
|  | B3-IT | 2269 | 8.6 | 0.137 | 3.29 | 9 | 0.004 | 3.82 | 7 |
|  | B4-IT | 14009 | 12.68 | 0.096 | 3.42 | 9 | 0.001 | 4.02 | 6 |
|  | B5-IT | 13403 | 11.72 | 0.088 | 3.60 | 19 | 0.001 | 4.12 | 7 |
| Croatian | B1-HR | 4155 | 7.48 | 0.099 | 3.58 | 10 | 0.002 | 4.36 | 8 |
|  | B2-HR | 12610 | 8.46 | 0.069 | 3.67 | 11 | 0.001 | 4.67 | 8 |
|  | B3-HR | 2970 | 6.46 | 0.098 | 3.54 | 9 | 0.003 | 4.47 | 8 |
|  | B4-HR | 15256 | 10.8 | 0.103 | 3.49 | 10 | 0.001 | 4.31 | 7 |
|  | B5-HR | 15985 | 9.82 | 0.077 | 3.77 | 22 | 0.001 | 4.49 | 8 |

Further analysis of B5 revealed a proportion of Latin and German, where Latin names, were declined in Croatian, and subsequently not lemmatized, which caused additional anomalies in the results. The English lemmatizer failed due to the same problem too.

Table 4 presents network measures for the B5 after the removal of Latin and German words. Compared to the initial B5 results from Tables 2 and 3 the $D_{DIR}$ and $D_{UNDIR}$ has decreased as expected. The undirected network had changed more than the directed. The results suggest that the Latin and German parts from the book created loops which caused $C_{DIR}$ to decrease. At the same time B5 in Italian behaves differently due to the close nature of Italian and Latin, which was partially captured during lemmatization.

**Table 4. New values for the directed and undirected networks of B5 by language.**

|       | $N$   | $<k>$ | $C_{DIR}$ | $L_{DIR}$ | $D_{DIR}$ |
|-------|-------|-------|-----------|-----------|-----------|
| B5-EN | 9355  | 6.754 | 0.054     | 3.59      | 13        |
| B5-IT | 10674 | 6.739 | 0.051     | 3.53      | 13        |
| B5-HR | 12817 | 5.463 | 0.042     | 3.82      | 14        |
|       |       |       | $C_{UNDIR}$ | $L_{UNDIR}$ | $D_{UNDIR}$ |
| B5-EN | 9355  | 62754 | 0.108     | 3.42      | 17        |
| B5-IT | 10674 | 6.739 | 0.103     | 3.43      | 15        |
| B5-HR | 12817 | 5.463 | 0.085     | 3.54      | 15        |

The differences across languages are presented in Fig. 1: in general, English has a higher clustering coefficient than Croatian.

Figure 1. Values of average degree and clustering coefficient for 15 directed networks grouped in languages.

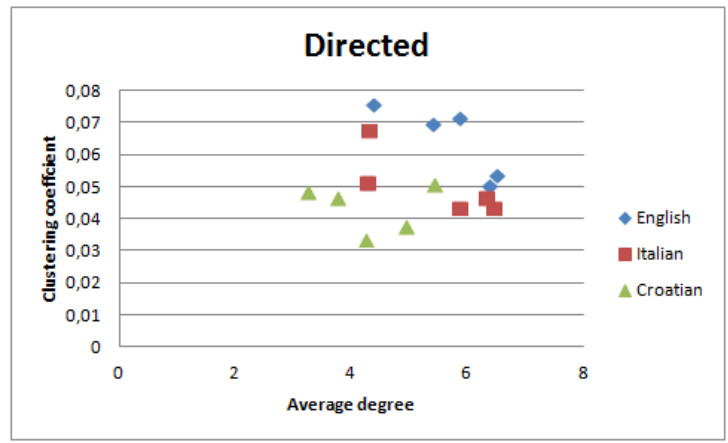

Average path lengths $L$ are the highest for Croatian, in the middle for Italian and the lowest for the English language networks as shown in Fig. 2. Similar results are presented in [12] where it is shown that Croatian language has larger values of $L$ and $D$ but $C$ twice as small than that of English. According to the graphs shown in Fig. 2 the average path length seems to be more influenced by the language than diameter. $D$ depends on the book size, but it is also sensitive to potential anomalies in the book's language, as is previously shown for book 5.

Figure 2. In the first row the ratio between the diameter of the books by language for directed and undirected networks is shown. The second row is the differentiation by the average path length.

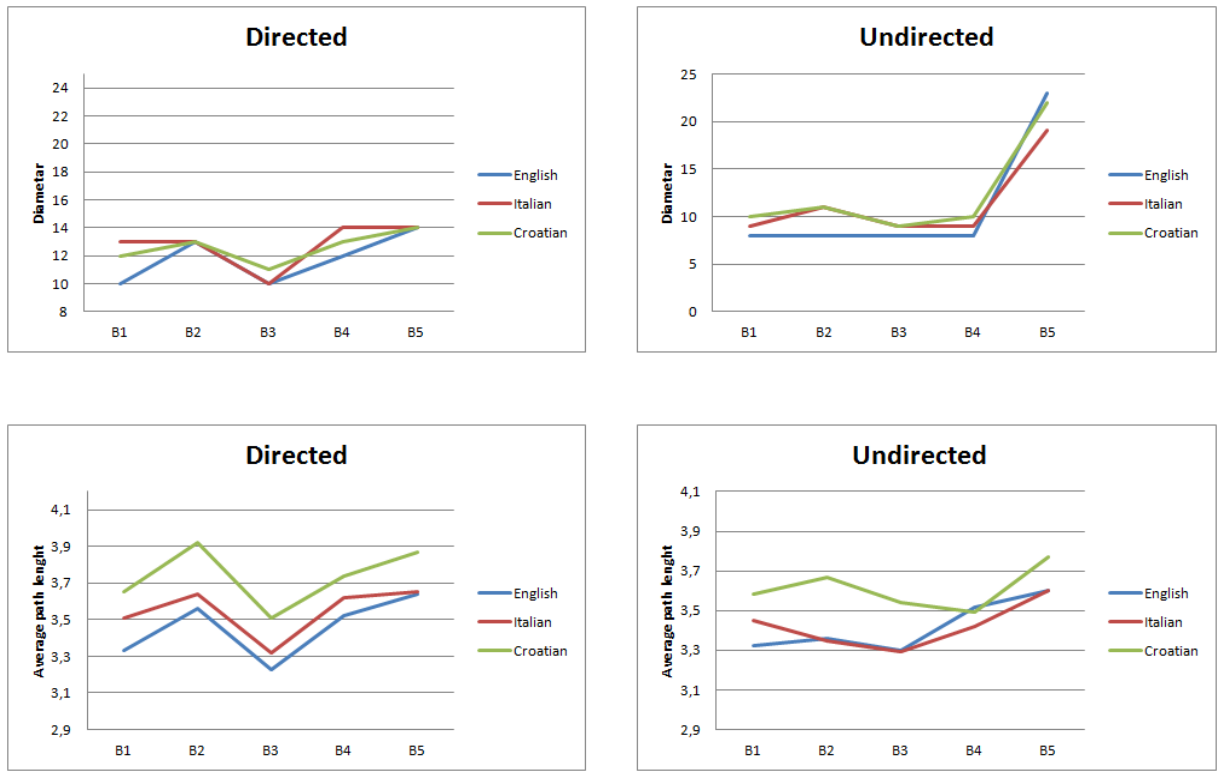

## 5. Conclusion

In this paper we have examined linguistic networks for Croatian, English and Italian language. The measures of 30 co-occurrences in directed and undirected networks for five books in three languages have been compared.

It has been shown that for all languages co-occurrence networks share small-world properties and corpus-sensitivity. Corpus size and possible anomalies in the text have an impact on the network structure in all three languages. An anomaly, such as the introduction of another language causes that diameter of undirected network becomes much higher than the diameter of a directed network as has been shown in the case of book B5. In addition, the results show that there are expected differences between the measurements for directed and undirected networks for all three languages.

However, further examination of the measurements of networks differs across languages: the clustering coefficient of English and Italian books is closer than that of Croatia. The Croatian language exhibits a higher path length in both directed and undirected networks, which can be caused by the relatively free word order. The word order of English is more precise than the Italian which is reflected in the directed networks in Fig. 2. The Croatian language also has the smallest clustering coefficient which can indicate to a richer language morphology. This result is partly sensitive to the degraded lemmatization of Croatian, which is also grounded in its complex morphology.

Finally, the average path length and clustering coefficient show language differentiation potential and should be analysed on larger corpora to test if they may be used as language classifiers. On the other hand the diameter is more related to books, which implies that it could be used as measure of the authors' vocabulary or verbosity. In further work all results should be tested on larger corpora in more languages in order to classify authorial or book genres from network parameters.

## 6. References

[1] R.F.I Cancho and R.V. Solé, The small- world of human language, Proceedings of The Royal Society of London. Series B(268), pages 2261–2265, 2001.
[2] R.V. Solé, B.C. Murtra, S. Valverde, L. Steels, Language Networks: their structure, function and evolution, Trends in Cognitive Sciences, 2005.
[3] R.F.I Cancho, R.V. Solé and R.Kohler, Patterns in syntactic dependency networks, Physical Review, E 69, 051915, 2004.
[4] H. Liu and C. Hu, Can syntactic networks indicate morphological complexity of a language, *EPL* **93** 28005, 2001.
[5] J. Borge-Holthoefer and A.Arenas, Semantic Networks: Structure and Dynamics, Entropy, 12, 1264-1302, 2010.
[6] S. Arbesman, S.H. Strogatz, M.S.Vitevitch, Comparative Analysis of Networks of Phonologically Similar Words in English and Spanish, Entropy, 12, pages 327-337, 2010.
[7] S. Arbesman, S.H. Strogatz, M.S.Vitevitch,The Structure of Phonological Networks across Multiple Languages, International Journal of Bifurcation and Chaos, 20(2):679-685, 2009.
[8] A.P. Masucci and G.J. Rodgers, Network properties of written human language, Physical Review E, 74.026102, 2006.
[9] F. C. Pembe and H. Bingol, Complex Networks in Different Languages: A Study of an Emergent Multilingual Encyclopedia, Proceedings of the Sixth International Conference on Complex Systems, 3, pages 612-617, 2008.
[10] L. Sheng and C. Li, English and Chinese language as weighted networks, Physica A, 388:2561-2570, 2009.
[11] H. Liu C. Jin, Language Clustering with Cord Co-occurrence Networks Based on Parallel Texts, Chin. Sc. Bull., 58(10), pages 1139-1144, 2013.
[12] A. A. Hagberg, D. A. Schult and P. J. Swart, Exploring network structure, dynamics, and function using NetworkX, in Proceedings of the 7th Python in Science Conference (SciPy2008), Gäel Varoquaux, Travis Vaught, and Jarrod Millman (Eds), (Pasadena, CA USA), pages 11–15, 2008.
[13] M. Bastian, S. Heymann, and M. Jacomy, Gephi: An open source software for exploring and manipulating networks, 2009.
[14] M. Medeiros Soares, G.Corso, L. S. Lucena, The Network of syllables in Portuguese, Physica A, 355(2-4): 678-684, 2005.
[15] K. Ban, I. Ivakić and A. Meštrović, A preliminary study of Croatian Language Syllable Networks, Mipro SP, pages 1697-1701, 2013.
[16] A. Mehri, A. H. Darooneh and A. Shariati, The complex networks approach for authorship attribution of books, Physica A, 391(7):2429–2437, 2012.

[17] S. Bird, E. Klein and E. Loper, Natural Language Processing with Python, O'Reilly Media, 2009.

[18] M. Bastian, S. Heymann, and M. Jacomy, Gephi: An open source software for exploring and manipulating networks, ICWSM, The AAAI Press, 2009.

[19] The Center for Information and Language Processing, TreeTagger - a language independent part-of-speech tagger, http://www.ims.uni-stuttgart.de/projekte/corplex/TreeTagger/, downloaded: June, 2013.

[20] M. Tadic and S. Fulgosi, Building the Croatian Morphological Lexicon, Proceedings of the EACL2003, pages :41-46, 2003.

[21] L. Antiqueira, T. A. S. Pardo, M. das G. V. Nunesand, O. N. Oliveira Jr., Inteligencia Artificial, 11(36), pages 51-58, 2007.

[22] P. Erdös, A. Rényi, On the evolution of random graphs, Publ. Math. Inst. Hung. Acad. Sci. 5 (1960) 17-60.